\def\year{2020}\relax
\documentclass[letterpaper]{article} 
\usepackage{aaai20}  
\usepackage{times}  
\usepackage{helvet} 
\usepackage{courier}  
\usepackage[hyphens]{url}  
\usepackage{graphicx} 
\usepackage{graphicx}  
\usepackage{amsmath}
\usepackage{booktabs}
\usepackage{algorithm} 
\usepackage{algorithmicx}  
\usepackage{algpseudocode} 
\usepackage{amssymb}
\newcommand{\citet}[1]{\citeauthor{#1} \shortcite{#1}} \newcommand{\citep}{\cite} 

\title{Generative Adversarial Zero-Shot Relational Learning \\
for Knowledge Graphs}
\author{
Pengda Qin,\textsuperscript{\rm 1}
Xin Wang,\textsuperscript{\rm 2}
Wenhu Chen,\textsuperscript{\rm 2}
Chunyun Zhang,\textsuperscript{\rm 3}
Weiran Xu\textsuperscript{\rm 1}
William Yang Wang\textsuperscript{\rm 2}\\
\textsuperscript{\rm 1}Beijing University of Posts and Telecommunications, China\\
\textsuperscript{\rm 2}University of California, Santa Barbara, USA\\
\textsuperscript{\rm 3}Shandong University of Finance and Economics, China\\
qinpengda@bupt.edu.cn, xwang@cs.ucsb.edu, wenhuchen@cs.ucsb.edu\\
zhangchunyun1009@126.com, xuweiran@bupt.edu.cn, william@cs.ucsb.edu
}
 
\begin{document}

\maketitle

\begin{abstract}
Large-scale knowledge graphs (KGs) are shown to become more important in current information systems. To expand the coverage of KGs, previous studies on knowledge graph completion need to collect adequate training instances for newly-added relations. In this paper, we consider a novel formulation, zero-shot learning, to free this cumbersome curation. For newly-added relations, we attempt to learn their semantic features from their text descriptions and hence recognize the facts of unseen relations with no examples being seen. For this purpose, we leverage Generative Adversarial Networks (GANs) to establish the connection between text and knowledge graph domain: The generator learns to generate the reasonable relation embeddings merely with noisy text descriptions. Under this setting, zero-shot learning is naturally converted to a traditional supervised classification task. Empirically, our method is model-agnostic that could be potentially applied to any version of KG embeddings, and consistently yields performance improvements on NELL and Wiki dataset.
\end{abstract}

\section{Introduction}
\label{sec:introduction}
Large-scale knowledge graphs collect an increasing amount of structured data, where nodes correspond to \emph{entities} and \emph{edges} reflect the relationships between head and tail entities. This graph-structured knowledge base has become a resource of enormous value, with potential applications such as search engine, recommendation systems and question answering systems. 
However, it is still incomplete and cannot cater to the increasing need of intelligent systems. To solve this problem, many studies \citep{bordes2013translating,trouillon2016complex} achieve notable performance on automatically finding and filling the missing facts of existing relations. But for newly-added relations, there is still a non-negligible limitation, and obtaining adequate training instances for every new relation is an increasingly impractical solution. 
Therefore, people prefer an automatic completion solution, or even a more radical method that recognizes unseen classes without seeing any training instances.

\begin{figure}[t]
\begin{center}
\includegraphics[width=0.95\columnwidth]{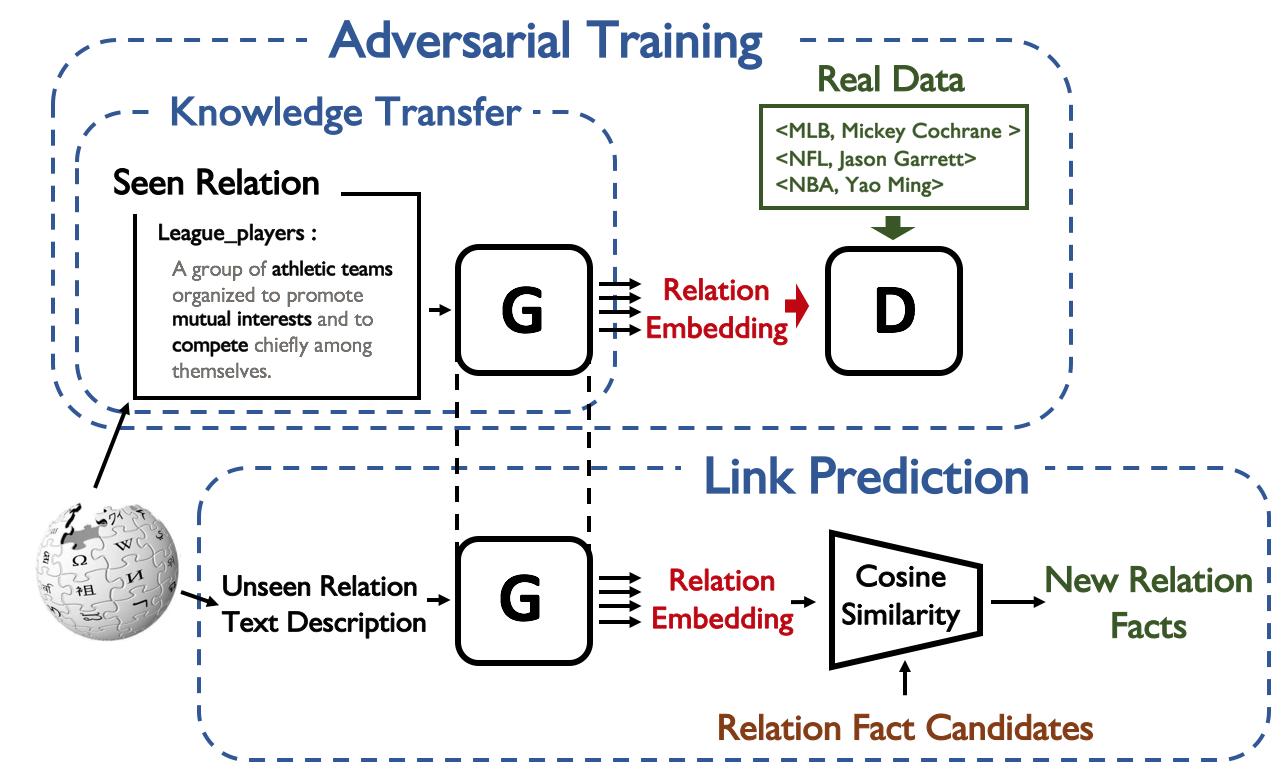}
\caption{Overview of our proposed approach. Through the adversarial training between generator (\textbf{G}) and discriminator (\textbf{D}), 
we leverage \textbf{G} to generate reasonable embeddings for unseen relations and predict new relation facts in a supervised way.}
\label{fig:display}
\end{center}
\end{figure}

Zero-shot learning aims to recognize objects or facts of new classes (\emph{unseen classes}) with no examples being seen during the training stage. Correspondingly, an appealing characteristic of human learning is that, with a certain accumulation of knowledge, people are able to recognize new categories merely from their text descriptions. 
Therefore, instead of learning from instances, the semantic features of new classes can be reflected by their textual descriptions. Moreover, textual descriptions contain rich and unambiguous information and can be easily accessed from dictionaries, encyclopedia articles or various online resources, which is critical for large-scale recognition tasks.

In this paper, we propose a zero-shot relational learning method for knowledge graph. As shown in Figure \ref{fig:display}, we convert zero-shot learning into a knowledge transfer problem. We focus on how to generate reasonable relation embeddings for unseen relations merely from their text descriptions.  Once trained, this system is capable of generating relation embeddings for arbitrary relations without fine-tuning. With these relation embeddings, the facts of unseen relations can be recognized simply by cosine similarity. To meet these requirements, the first challenge is how to establish an effective knowledge transfer process from text semantic space to knowledge graph semantic space. We leverage the conditional GANs to generate the plausible relation embeddings from text descriptions and provide the inter-class diversity for unseen relations. The second challenge is the noise suppression of text descriptions. Human language expression always includes irrelevant words (such as function words) for identifying target relations. As in Figure \ref{fig:display}, the bold words are more critical for the meaning of relation \emph{League\_players}; Therefore, the indiscriminate weights for words will lead to inferior performance. For this problem, We adopt the simple bag-of-words model based on word embeddings; Simultaneously, we calculate the TF-IDF features to down-weight the importance of the less relevant words for zero-shot learning. Our main contributions are three-fold:
\begin{itemize}
\item We are the first to consider zero-shot learning for knowledge graph completion, and propose a generative adversarial framework to generate reasonable relation embeddings for unseen relations merely from text descriptions;
\item Our method is model-agnostic and can be potentially applied to any version of KG embeddings;
\item We present two newly constructed datasets for zero-shot knowledge graph completion and show that our method achieves better performance than various embedding-based methods.
\end{itemize}

\section{Related Work}
\label{sec:related_work}
Currently, representation learning \citep{nickel2011three} has been the widely-used way to model knowledge graph information. TransE \citep{bordes2013translating} projects relations and entities from symbolic space to vector space, and the missing links of the existing relations can be inferred via simple vector operations. Subsequently, many notable embedding-based studies \citep{yang2014embedding,trouillon2016complex} are proposed for knowledge graph completion. However, these methods are incapable of any action when dealing with newly-add relations. Unlike that, the proposed method still has good recognition ability for the relation facts of unseen relations. \citet{xiong2018one} proposes a few-shot learning method that learns a matching network and predicts the unseen relation facts by calculating their matching score with a few labeled instances. In contrast, our method follows the zero-shot setting and do not need any training instances for unseen relations. KBGAN \citep{cai2018kbgan} adopts adversarial training to learn a better discriminator via selecting high-quality negative samples, but it still focuses on the link prediction of existing relations.

The core of zero-shot learning (ZSL) is realizing knowledge sharing and inductive transfer between the seen and the unseen classes, and the common solution is to find an intermediate semantic representation. For this purpose, \citet{akata2013label} propose an \textbf{attribute-based model} that learns a transformation matrix to build the correlations between attributes and instances. However, attribute-based methods still depend on lots of human labor to create attributes, and are sensitive to the quality of attributes. \textbf{Text-based methods} \citep{qiao2016less} are to create the intermediate semantic representation directly from the available online unstructured text information. To suppress the noise in raw text, \citet{wang2019learning} leverage TF-IDF features to down-weight the irrelevant words. 
As for model selection, the ZSL framework of \citet{zhu2018generative} greatly inspires us, which leverages a conditional GANs model to realize zero-shot learning on image classification task.
Currently, the majority of ZSL research works are from the computer vision domain. In the field of Natural Language Processing, \citet{artetxe2019massively} use a single sentence encoder to finish the multilingual tasks by only training the target model on a single language. To the best of our knowledge, this work is the first zero-shot relational learning for knowledge graphs.

\section{Background}
\label{sec:background}
\subsection{Zero-Shot Learning Settings}
Here we present the problem definition and some notations of zero-shot learning based on knowledge graph completion task. Knowledge graph is a directed graph-structured knowledge base and constructed from tremendous relation fact triples $\{(e_1, r, e_2)\}$. Since the proposed work aims to explore the recognition ability when meeting the newly-added relations, our target can be formulated as predicting the tail entity $e_2$ given the head entity $e_1$ and the query relation $r$. To be more specific, for each query tuple $(e_1, r)$, there are a ground-truth tail entity $e_2$ and a candidate set $C_{(e_1, r)}$; our model needs to assign the highest ranking to $e_2$ against the rest candidate entities $e_2^{\prime} \in C_{(e_1, r)}$.
According to the zero-shot setting, there are two different relation sets, the seen relation set $R_s = \{r_s\}$ and the unseen relation set $R_u = \{r_u\}$, and obviously $R_{s} \cap R_{u} = \emptyset$. 

At the start, we have a background knowledge graph $\mathcal{G}$ that collects a large scale of triples $\mathcal{G} = \{(e_1, r_s, e_2)|e_1 \in E, r_s \in R_s, e_2 \in E \}$, and $\mathcal{G}$ is available during the zero-shot training stage. With this knowledge graph, we establish a training set $D_s = \{(e_1, r_s, e_2, C_{(e_1, r_s)})\}$ for the seen relations $r_s \in R_s$. During testing, the proposed model is to predict the relation facts of unseen relations $r_u \in R_u$. As for textual description, we automatically extract an online textual description $T$ for each relation in $R_{s} \cup R_{u}$. In view of feasibility, we only consider a closed set of entities; More specifically, each entity that appears in the testing triples is still in the entity set $E$. Thus, our testing set can be formulated as $D_u = \{(e_1, r_u, e_2, C_{(e_1, r_u)})|e_1 \in E, r_u \in R_u, e_2 \in E\}$. With the same requirement of the training process, the ground-truth tail entity $e_2$ needs to be correctly recognized by ranking $e_2$ with the candidate tail entities $e_2^{\prime} \in C_{(e_1, r_u)}$. We leave out a subset of $D_s$ as the validation set $D_{valid}$ by removing all training instances of the validation relations.

\subsection{Generative Adversarial Models}
Generative adversarial networks  \citep{goodfellow2014generative} have been enjoying the considerable success of generating realistic objective, especially on image domain. 
The generator aims to synthesize the reasonable pseudo data from random variables, and the discriminator is to distinguish them from the real-world data. Besides random variables, \citet{zhang2017stackgan} and \citet{zhu2018generative} have proved that the generator possesses the capability of knowledge transfer from the textual inputs. The desired solution of this game is Nash equilibrium; Otherwise, it is prone to unstable training behavior and mode collapse. Recently, many works \citep{arjovsky2017wasserstein,heusel2017gans} have been proposed to effectively alleviate this problem. Compared with the non-saturating GAN\footnote{Goodfellow's team \citep{fedus2017many} clarified that the standard GAN \citep{goodfellow2014generative} should be uniformly called non-saturating GANs.} \citep{goodfellow2014generative},  WGAN \citep{arjovsky2017wasserstein} optimizes the original objective by utilizing \textbf{Wasserstein distance} between real and fake distributions.
On this basis, \citet{gulrajani2017improved} propose a \textbf{gradient penalty} strategy as the alternative to the weight clipping strategy of WGAN, in which way to better satisfy Lipschitz constraint. \citet{miyato2018spectral} introduce \textbf{spectral normalization} to further stabilize the training of discriminator. Practice proves that our model benefits a lot from these advanced strategies. 

Besides, because KG triples are from different relations, our task should be regarded as a class conditional generation problem, and it is a common phenomenon in real-world datasets. 
\textbf{ACGAN} \citep{odena2017conditional} adds an auxiliary category recognition branch to the cost function of the discriminator and apparently improves the diversity of the generated samples\footnote{\citet{miyato2018cgans} proposes a projection-based way to alleviate model collapse when dealing with too many classes, but it is not suitable for our margin ranking loss.}. 
Spectral normalization is also impressively beneficial to the diversity of the synthetic data.

\section{Methodology}
\label{sec:methodology}
\begin{figure*}[t]
\begin{center}
\includegraphics[width=1.8\columnwidth]{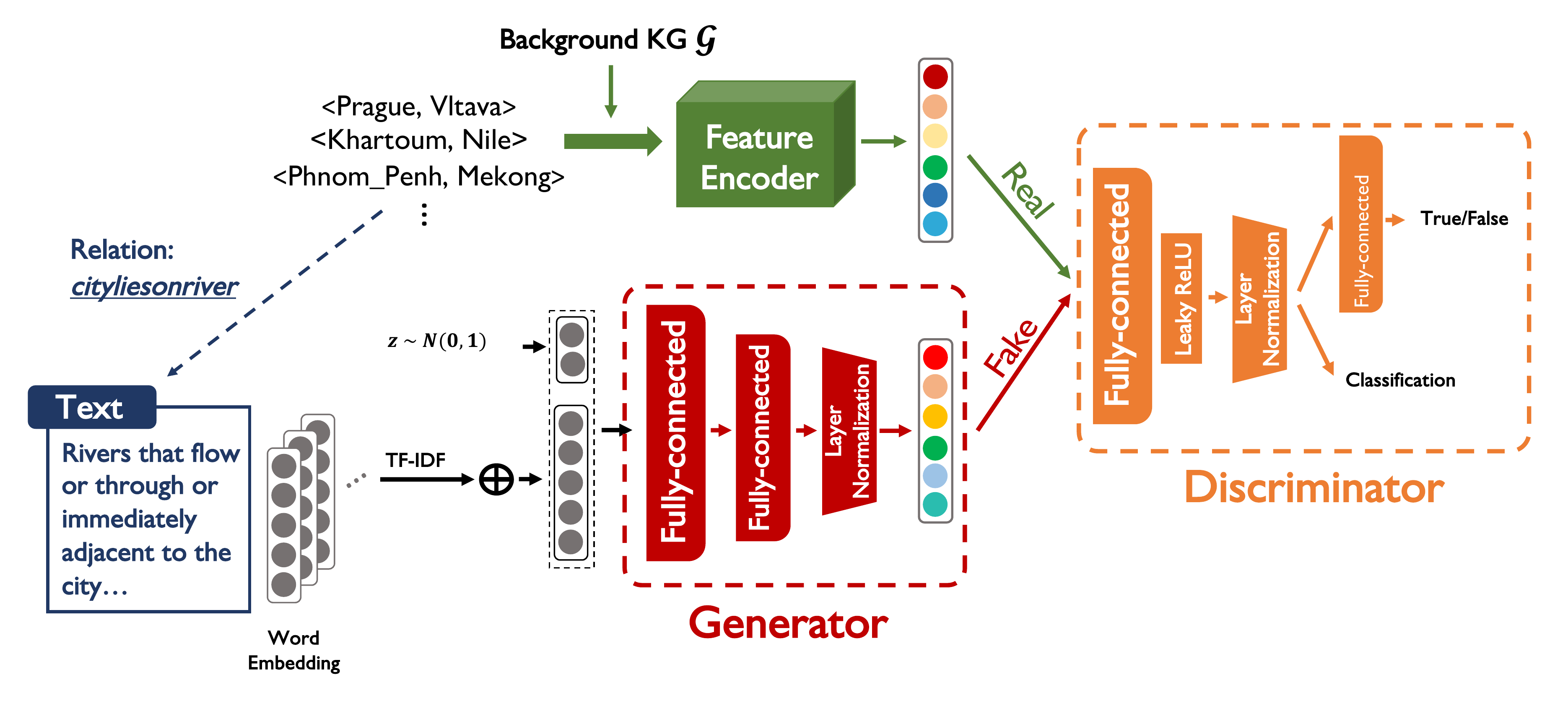}
\caption{Overview of the proposed generative model for zero-shot knowledge graph relational learning. Firstly, entity pairs of KG relation facts are fed into the \textbf{feature encoder} to calculate their semantic representations (\emph{real data}). Then, the \textbf{Generator} aims to generate the relation embeddings (\emph{fake data}) from the denoised text representation and random vector $z$. Finally, the \textbf{Discriminator} is designed to distinguish real data from fake data and assign the correct relation types to them.}
\label{fig:KG-GANs}
\end{center}
\end{figure*}

In this section, we describe the proposed model for zero-shot knowledge graph relational learning. As shown in Figure~\ref{fig:KG-GANs}, the core of our approach is the design of a conditional generative model to learn the qualified relation embeddings from raw text descriptions. Fed with text representations, the \textbf{generator} is to generate the reasonable relation embeddings that reflect the corresponding relational semantic information in the knowledge graph feature space. Based on this, the prediction of unseen relations is converted to a simple supervised classification task. On the contrary, the \textbf{discriminator} seeks to separate the fake data from the real data distribution and identifies the relation type as well. For real data representations, it is worth mentioning that we utilize a \textbf{feature encoder} to generate reasonable real data distribution from KG embeddings. The feature encoder is trained in advance from the training set and fixed during the adversarial training process.

\begin{figure}[t]
\begin{center}
\includegraphics[width=1.0\columnwidth]{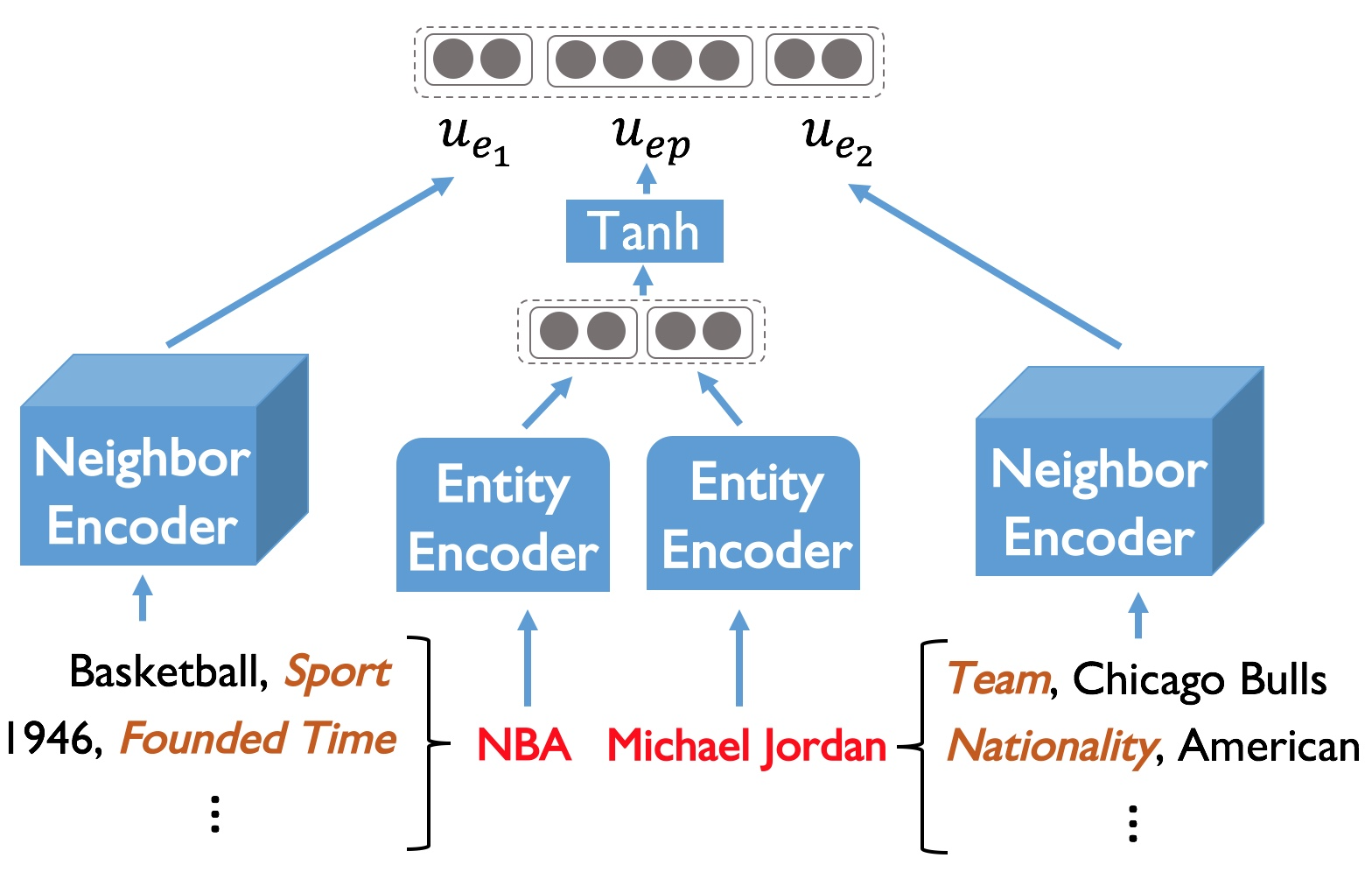}
\caption{Framework of Feature Encoder. For entity pair (\emph{NBA}, \emph{Michael Jordan}), \textbf{neighbor encoder} models the one-hop graph-structured information, and \textbf{entity encoder} extracts the useful information from entitie pairs themselves.}
\label{fig:feature_encoder}
\end{center}
\end{figure}

\subsection{Feature Encoder}
Traditional KG embeddings fit well on the seen relation facts during training; However, the optimal zero-shot feature representations should provide the cluster-structure distribution for both seen and unseen relation facts. Therefore, we design a feature encoder to learn better data distribution from the pretrained KG embeddings and one-hop structures.

\noindent \textbf{Network Architecture:} Feature encoder consists of two sub-encoders, the neighbor encoder and the entity encoder. 
In the premise of the feasibility of real-world large-scale KGs, for each entity $e$, we only consider the one-hop neighbors $\mathcal{N}_e = \{(r^n, e^{n})|(e,r^n,e^{n}) \in \mathcal{G}\}$ \citep{xiong2018one}. 
Therefore, we adopt the neighbor encoder to generate structural representations. 
Given a KG embedding matrix of dimension $d$, we first utilize an embedding layer to look up the corresponding neighbor entity and relation embeddings $v_{e^n}$, $v_{r^n}$. Then, the structure-based representation $u_e$ of entity $e$ is calculated \citep{schlichtkrull2018modeling} as below,
\begin{equation}
\begin{split}
&\mathnormal{f}_1 (v_{r^n}, v_{e^{n}}) = W_1 (v_{r^n} \oplus v_{e^{n}}) + b_1\\
&{u_e} = \sigma(\frac{1}{\vert \mathcal{N}_e \vert} \sum_{(r^n,e^{n})\in \mathcal{N}_e} \mathnormal{f}_1 (v_{r^n}, v_{e^{n}})),
\label{equ:neighbor_encoder}
\end{split}
\end{equation}
where $\sigma$ is \emph{tanh} activation function, and $\oplus$ denotes the concatenation operation. In consideration of scalability, we set an upper limit for the number of neighbors. Besides, we also apply a simple feed-forward layer as the entity encoder to extract the information from entity pair $(e_1, e_2)$ themselves,
\begin{equation}
\begin{split}
&\mathnormal{f}_2 (v_e) = W_2 (v_e) + b_2\\
&{u_{ep}} = \sigma(\mathnormal{f}_2 (v_{e_1}) \oplus \mathnormal{f}_2 (v_{e_2})).
\label{equ:entity_encoder}
\end{split}
\end{equation}

To sum up, as Figure~\ref{fig:feature_encoder}, the relation fact representation is formulated as the concatenation of the neighbor embeddings $u_{e_1}$, $u_{e_2}$ and the entity pair embedding $u_{ep}$,
\begin{equation}
x_{(e_1, e_2)} = u_{e_1} \oplus u_{ep} \oplus u_{e_2},
\label{equ:fact_embed}
\end{equation}
where $W_1 \in \mathnormal{R}^{d \times 2d}$, $W_2 \in \mathnormal{R}^{d \times d}$, $b_1, b_2 \in \mathnormal{R}^d$ are the learned parameters.\\

\noindent \textbf{Pretraining Strategy:}
The core of this pretraining step is to learn the cluster-structure data distribution that reflects a higher intra-class similarity and relatively lower inter-class similarity. 
The traditional supervised way with cross-entropy loss gives inter classes too much penalty and is impracticable for unseen classes. Thus, we adopt an effective matching-based way via margin ranking loss \citep{xian2017zero}. For each relation $r_s$, in one training step, we first randomly take out $k$ reference triples $\{e_1^{\star}, r_s, e_2^{\star}\}$ from the training set, a batch of positive triples $\{e_1^{+}, r_s, e_2^{+}\}$ from the rest of training set, and a batch of negative triples $\{e_1^{+}, r_s, e_2^{-}\}$\footnote{The negative triples are generated by polluting the tail entities.}. Then we use the feature encoder to generate the reference embedding  $x_{(e_1^{\star},e_2^{\star})}$, and calculate its cosine similarity respectively with $x_{(e_1^{+},e_2^{+})}$ and $x_{(e_1^{-},e_2^{-})}$ as $\mathnormal{score}_{\omega}^{+}$ and $\mathnormal{score}_{\omega}^{-}$. Therefore, the margin ranking loss can be described as below,
\begin{equation}
\mathnormal{L}_{\omega} = \mathnormal{max}(0, \gamma + \mathnormal{score}_{\omega}^{+} -  \mathnormal{score}_{\omega}^{-}),
\label{equ:ranking_loss}
\end{equation}
where $\omega = \{W_1, W_2, b_1, b_2\}$ is the parameter set to learn and $\gamma$ denotes the margin.
The best parameters of the feature encoder are determined by the validation set $D_{valid}$.

\subsection{Generative Adversarial Model}

\noindent \textbf{Generator:}
The generator is to generate the plausible relation embeddings from textual descriptions. First, for text representations, we simply adopt the bag-of-words method, where words are encoded with the pretrained word embeddings \citep{mikolov2013distributed,pennington2014glove}
as in Figure \ref{fig:KG-GANs}. To suppress the noise information, we first remove stop-words and punctuations, and then evaluate the importance of the rest words via TF-IDF features \citep{salton1988term}. Thus, the text embedding $T_r$ is the vector sum of word embeddings weighted by TF-IDF values. To meet the GANs requirements, we concatenate each text embedding with a random vector $z \in \mathnormal{R}^Z$ sampled from Gaussian distribution $N(0,1)$. As in Figure~\ref{fig:KG-GANs}, the following knowledge transfer process is completed by two fully-connected (FC) layers and a layer normalization operation. So, relation embedding $\tilde{x}_r$ is generated by the generator $\tilde{x}_r \gets G_\theta (T_r, z)$ with parameters $\theta$. To avoid mode collapse and improve diversity, we adopt the Wasserstein loss and an additional classification loss. This classification loss is formulated as the margin ranking loss as equation~\ref{equ:ranking_loss}. Here, the cluster center $x_c^r = \frac{1}{N_r} \sum_{i=1}^{N_r} x_{(e_1, e_2)}^i$ is regarded as the real relation representation, where $N_r$ is the number of facts of relation $r$. Thus, positive scores are calculated from $x_c^r$ and $\tilde{x}_r$; Negative scores are calculated from $x_c^r$ and negative fact representations where negative facts are generated by polluting tail entities.
In addtion, visual pivot regularization \citep{zhu2018generative} $\mathnormal{L}_{P}$ is also applied to provide enough inter-class discrimination.
\begin{equation}
\begin{split}
{\mathnormal{L}_{G_\theta}} = & -\mathbb{E}_{z \sim p_z}[D_{\phi}(G_\theta (T_r, z))]\\
&+ {\mathnormal{L}_{cls}}{(G_\theta (T_r, z))} + \mathnormal{L}_{P},
\label{equ:generator}
\end{split}
\end{equation}

\noindent \textbf{Discriminator:}
The discriminator attempts to distinguish whether an input is the real data $x_{(e_1, e_2)}$ or the fake one $\tilde{x}_r$; Besides, it also needs to correctly recognize their corresponding relation types. As in Figure~\ref{fig:KG-GANs}, the input features are first transformed via a FC layer with Leaky ReLU \citep{maas2013rectifier}. Following this, there are two network branches. The first branch is a FC layer that acts as a binary classifier to separate real data from fake data, and we utilize the Wasserstein loss as well. The other branch is the classification performance. In order to stabilize training behavior and eliminate mode collapse, we also adopt the \textbf{gradient penalty} $\mathnormal{L}_{GP}$ to enforce the Lipschitz constraint. It penalizes the model if the gradient norm moves away from its target norm value 1. 
In summary, the loss function of the discriminator is formulated as:
\begin{equation}
\begin{split}
\mathnormal{L}_{D_\phi} = &\mathbb{E}_{z \sim p_z}[D_{\phi}(G_\theta (T_r, z))] - \mathbb{E}_{x \sim p_{data}}[D_{\phi}(x)] \\
& + \frac{1}{2}{\mathnormal{L}_{cls}}{(G_\theta (T_r, z))} + \frac{1}{2}{\mathnormal{L}_{cls}}(x) + \mathnormal{L}_{GP}.
\label{equ:discriminator}
\end{split}
\end{equation}

\begin{algorithm}[t]  
        \caption{The proposed generative adversarial model for zero-shot knowledge graph relational learning.}  
        \begin{algorithmic}[1]  
            \Require The number of training steps $N_{step}$, the ratio of iteration time between $D$ and $G$ $[n_d:1]$, Adam hyperparameters $\alpha$, $\beta_1$, $\beta_2$
            \State Load the pre-trained feature encoder
            \State Initialize parameters $\theta$, $\phi$ for $G$, $D$
            \For{$i = 1 \to N_{step}$}
                \For{$i_d \to n_d$}
                	\State Sample a subset $R_s^D$ from $R_s$ and obtain text $T_{r_s}^D$, random noise $z$
                	\State Sample a minibatch of triples $B_D^+$ of $R_s^D$
                	\State $\tilde{x}_{r_s}^D \gets G_\theta (T_{r_s}^D, z)$
                	\State Obtain negative set $B_D^-$ for $B_D^+$ and $\tilde{x}_{r_s}^D$
                	\State Compute the loss of $D$ using Eq.~\ref{equ:discriminator}
                	\State $\phi \gets Adam(\nabla_{\phi} \mathnormal{L}_D, \phi, \alpha, \beta_1, \beta_2)$
                \EndFor
                \State Sample a subset $R_s^G$ from $R_s$ and obtain text $T_{r_s}^G$, random noise $z$
                \State $\tilde{x}_{r_s}^G \gets G_\theta (T_{r_s}^G, z)$ 
                \State Obtain negative set $B_G^-$ for $\tilde{x}_{r_s}^G$
                \State Compute the loss of $G$ using Eq.~\ref{equ:generator}
                \State $\theta \gets Adam(\nabla_{\theta} \mathnormal{L}_G, \theta, \alpha, \beta_1, \beta_2)$
            \EndFor
        \end{algorithmic}  
    \end{algorithm}

\subsection{Predicting Unseen Relations}
After adversarial training, given a relation textual description $T_{r_u}$, the generator can generate its plausible relation embedding $\tilde{x}_{r_u} \gets G_\theta (T_{r_u}, z)$. 
For a query tuple $(e_1, r_u)$, the similarity ranking value $\mathnormal{score}_{(e1, r_u, e_2)}$ 
can be calculated by the cosine similarity between $\tilde{x}_{r_u}$ and $x_{(e_1, e_2)}$. It is worth mentioning that, since $z$ can be sampled indefinitely, we can generate an arbitrary number $N_{test}$ of generated relation embeddings ${\{\tilde{x}_{r_u}^i\}}_{i = 1,2,,N_{test}}$. For the better generalization ability, we utilize the average cosine similarity value as the ultimate ranking score,
\begin{equation}
\mathnormal{score}_{(e_1, r_u, e_2)} = \frac{1}{N_{test}} \sum_{i=1}^{N_{test}} \mathnormal{score}_{(e_1, r_u, e_2)}^i.
\label{equ:zero-shot}
\end{equation}

\section{Experiments}
\label{sec:experiments}

\subsection{Datasets and Evaluation Protocols}
\begin{table}[!htbp]
\centering
\begin{tabular}{ccccc}
\toprule
\bf{Dataset} & \bf{\# Ent.} & \bf{\# Triples} & \bf{\# Train/Dev/Test} \\
\midrule
\textbf{NELL-ZS} & 65,567 & 188,392 & 139/10/32 \\
\textbf{Wiki-ZS} & 605,812 & 724,967 & 469/20/48 \\
\bottomrule
\end{tabular}
\caption{\label{table:Dataset} Statistics of the constructed zero-shot datasets for KG link prediction. \# Ent. denotes the number of unique entities. \# Triples denotes the amount of relation triples. \# Train/Dev/Test denotes the number of relations for  training/validation/testing.}
\end{table}

\begin{table*}[t]
\centering
 \begin{tabular}{lcccc|cccc} \toprule
& \multicolumn{4}{c}{\textbf{NELL-ZS}} & \multicolumn{4}{c}{\textbf{Wiki-ZS}}\\ 
\cmidrule{2-5} \cmidrule{6-9}
Model & MRR & Hits@10 & Hits@5 & Hits@1 & MRR & Hits@10 & Hits@5 & Hits@1 \\ \midrule
ZS-TransE & 0.097 & 20.3 & 14.7 & 4.3 & 0.053 & 11.9 & 8.1 & 1.8\\
ZS-DistMult & 0.235 & 32.6 & 28.4 &  18.5 & 0.189 & 23.6 & 21.0 & 16.1\\
ZS-ComplEx & 0.216 & 31.6 & 26.7 & 16.0 & 0.118 & 18.0 & 14.4 & 8.3 \\
\midrule
{ZSGAN}$_{KG}$ (TransE) & 0.240 & \underline{\textbf{37.6}} & \underline{\textbf{31.6}} & 17.1 & 0.185 & 26.1 & 21.3 & 14.1\\
{ZSGAN}$_{KG}$ (DistMult) & \underline{\textbf{0.253}} & 37.1 & 30.5 & \underline{\textbf{19.4}} & \underline{\textbf{0.208}} & \underline{\textbf{29.4}} & \underline{\textbf{24.1}} & \underline{\textbf{16.5}}\\
{ZSGAN}$_{KG}$ (ComplEx-re) & 0.231 & 36.1 & 29.3 & 16.1 & 0.186 & 25.7 & 21.5 & 14.5\\
{ZSGAN}$_{KG}$ (ComplEx-im) & 0.228 & 32.1 & 27.0 & 17.4 & 0.185 & 24.8 & 20.9 & 14.7\\
\bottomrule
 \end{tabular}
 \caption{Zero-shot link prediction results on the unseen relations. 
 The proposed baselines are shown at the top of the table; Our generative adversarial model is denoted as {ZSGAN}$_{KG}$ and the results are shown at the bottom. 
 \textbf{Bold} numbers denote the best results, and \underline{Underline} numbers denote the best ones among our {ZSGAN}$_{KG}$ methods.}
 \label{table:results}
\end{table*}

\noindent \textbf{KG Triples:} Because there is not available zero-shot relational learning dataset for knowledge graph, we decide to construct two reasonable datasets from the existing KG Datasets. We select NELL\footnote{\url{http://rtw.ml.cmu.edu/rtw/}} \citep{carlson2010toward} and Wikidata\footnote{\url{https://pypi.org/project/Wikidata/}} for two reasons: the large scale and the existence of official relation descriptions.
For NELL, we take the latest dump and remove those inverse relations.
The dataset statistics are presented in Table ~\ref{table:Dataset}. 

\noindent \textbf{Textual Description:} The NELL and Wikidata are two well-configured knowledge graphs.
Our textual descriptions consist of multiple information.
For NELL, we integrate the relation description and its entity type descriptions. For Wikidata, each relation is represented as a property item. Besides its property description, we also leverage the attributes \emph{P31}, \emph{P1629}, \emph{P1855} as the additional descriptions. 

\noindent \textbf{Evaluation Protocols:} Following previous works \citep{yang2014embedding,xiong2018one}, we use two common metrics, mean reciprocal ranking (MRR) and hits at 10 (H@10), 5 (H@5), 1 (H@1).
During testing, candidate sets are constructed by using the entity type constraint \citep{toutanova2015representing}.

\subsection{Baselines}
In our experiments, the baselines include three commonly-used KG embedding methods: TransE \citep{bordes2013translating}, DistMult \citep{yang2014embedding} and ComplEx \citep{trouillon2016complex}. Obviously, these original models cannot handle zero-shot learning. Therefore, based on these three methods, we propose three zero-shot baselines, {\bf ZS-TransE}, {\bf ZS-DistMult} and {\bf ZS-ComplEx}. Instead of randomly initializing a relation embedding matrix to represent relations, we add a feed-forward network with the same structure\footnote{This feed-forward network does not receive random noise $z$ as input.} of our generator to calculate relation embeddings for these three methods. Equally, we utilize text embeddings as input and fine-tune this feed-forward network and entity embeddings via their original objectives. Under this setting, the unseen relation embeddings can be calculated via their text embeddings, and the unseen relation facts can be predicted via their original score functions. RESCAL \citep{nickel2011three} cannot directly adopt the same feed-forward network for zero-shot learning; For a fair comparison, we do not consider this KG embedding method.

\subsection{Implementation Details}
For NELL-ZS dataset, we set the embedding size as 100. For Wiki-ZS, we set the embedding size as 50 for faster training.
The three aforementioned baselines are implemented based on the Open-Source Knowledge Embedding toolkit \emph{OpenKE}\footnote{\url{https://github.com/thunlp/OpenKE}}\citep{han2018openke}, and their hyperparameters are tuned using the Hits@10 metric on the validation set $D_{valid}$.
The proposed generative method uses the pre-trained KG embeddings as input, which are trained on the triples in the training set. For TransE and DistMult, we directly use their 1-D vectors. For ComplEx, we set two experiments by respectively using the real embedding matrix and the imaginary embedding matrix as in Table \ref{table:results}.
For both the feature encoder and the generative model, we adopt the Adam \citep{kingma2014adam} for parameter updates, and the margin $\gamma$ is set as 10.0. For feature encoder, the upper limit of the neighbor number is 50, the number of reference triples $k$ in one training step is 30, and the learning rate is $5e^{-4}$. For the generative model, the learning rate is $1e^{-4}$, and 
$\beta_1$, $\beta_2$ are set as 0.5, 0.9 respectively. When updating the generator one time, the iteration number $n_d$ of the discriminator is 5. The dimension of the random vector $z$ is 15, and the number of the generated relation embedding $N_{test}$ is 20. Spectral normalization is applied for both generator and discriminator. These hyperparameters are also tuned on the validation set $D_{valid}$. 
As for word embeddings, we directly use the released word embedding set \emph{GoogleNews- vectors-negative300.bin}\footnote{\url{http://code.google.com/p/word2vec/}} of dimension 300.

\begin{table*}[!htbp]
    \centering
    \begin{tabular}{lcc|ll|ll}
    \toprule
    &  &  & \multicolumn{2}{c}{MRR} & \multicolumn{2}{c}{Hits@10} \\ 
\cmidrule{4-5} \cmidrule{6-7}
       Relations &  \# Can. Num. &  \# Cos. Sim. & {ZSGAN}$_{KG}$ & ZS-DistMult & {ZSGAN}$_{KG}$ & ZS-DistMult \\
    \midrule
       animalThatFeedOnInsect & 293 & 0.8580 & {\bf 0.347} & 0.302 & {\bf 63.4} & 61.8 \\
       automobileMakerDealersInState & 600 & 0.1714 & \textbf{0.066} & 0.039 & \textbf{15.4} & 5.1\\
       animalSuchAsInvertebrate & 786 & 0.7716 & \textbf{0.419} & 0.401 & \textbf{59.8} & 57.6\\
       sportFansInCountry & 2100 & 0.1931 & \textbf{0.066} & 0.007 & \textbf{15.4} & 1.3\\
       produceBy & 3174 & 0.6992 & \textbf{0.467} & 0.375 & \textbf{65.3} & 51.2\\
       politicalGroupOfPoliticianus & 6006 & 0.2211 & 0.018 & \textbf{0.039} & \textbf{5.3} & 3.9\\
       parentOfPerson & 9506 & 0.5836 & 0.343 & \textbf{0.381} & 56.2 & {\bf 60.4}\\
       teamCoach & 10569 & 0.6764 & \textbf{0.393} & 0.258 & \textbf{53.7} & 39.9\\
    \bottomrule
    \end{tabular}
    \caption{Quantitative analysis of the generated relation embeddings by our generator. These presented relations are from the NELL test relation set.  ``\# Can. Num." denotes the number of candidates of test relations. For one relation, ``\# Cos. Sim." denotes the mean cosine similarity between the corresponding generated relation embedding and the cluster center $x_c^r$ of the relation triples.}
    \label{table:per_relation}
\end{table*}

\subsection{Results}
Compared with baselines, the link prediction results of our method are shown in Table \ref{table:results}. Even though NELL-ZS and Wiki-ZS have different scales of triples and relation sets, the proposed generative method still achieves consistent improvements over various baselines on both two zero-shot datasets. It demonstrates that the generator successfully finds the intermediate semantic representation to bridge the gap between seen and unseen relations and generates reasonable relation embeddings for unseen relations merely from their text descriptions. Therefore, once trained, our model can be used to predict arbitrary newly-added relations without fine-tuning, which is significant for real-world knowledge graph completion.

\noindent \textbf{Model-Agnostic Property:} From the results of baselines, we can see that their performances are sensitive to the particular method of KG embeddings. Taking MRR and Hits@10 as examples, ZS-DistMult yeilds respectively 0.138 and 12.3\% higher performance than ZS-TransE on NELL-ZS dataset. However, our method achieves relatively consistent performance no matter which KG embedding matrix is used.

\begin{figure}[t]
\begin{center}
\includegraphics[width=1.0\columnwidth]{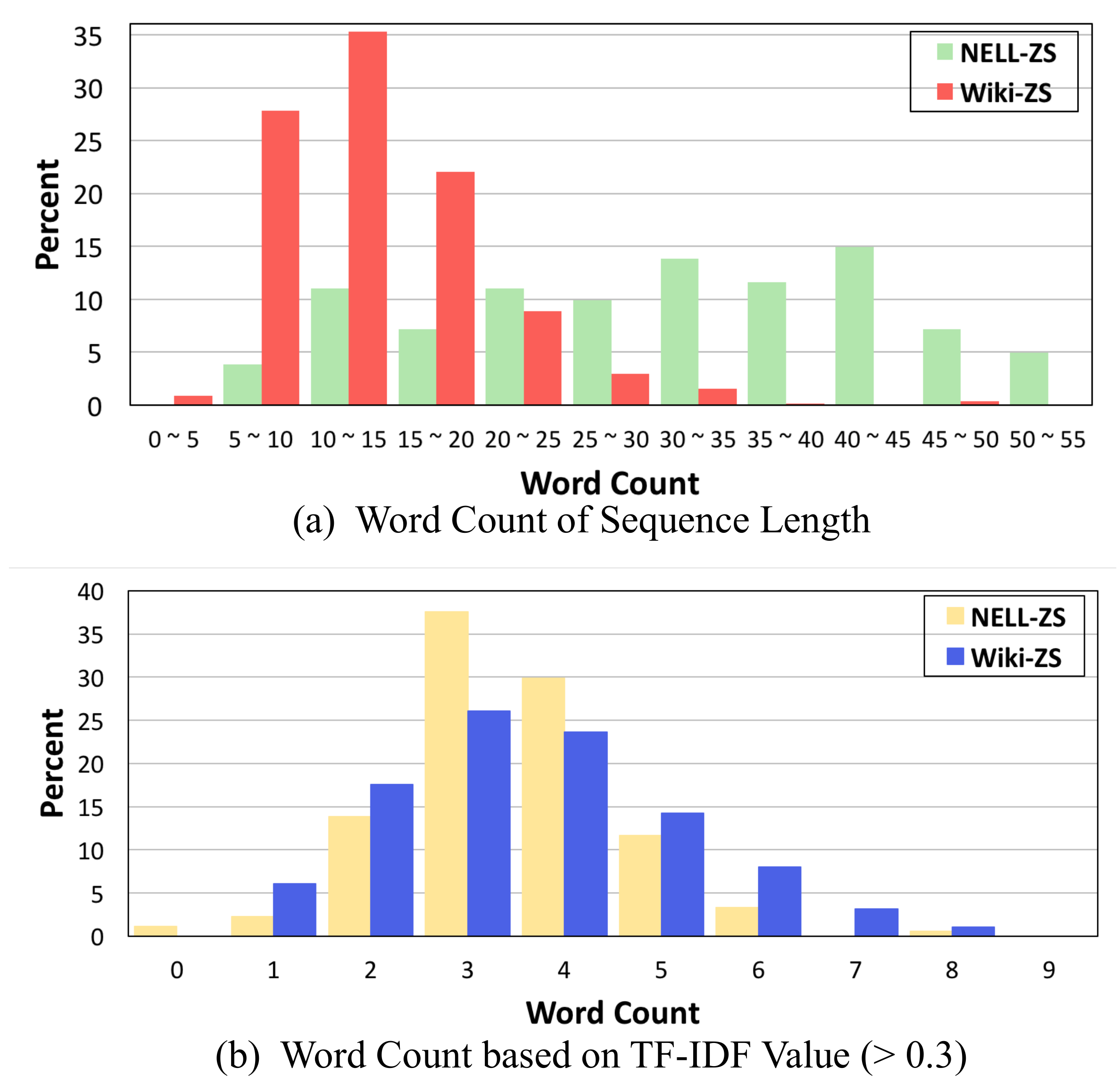}
\caption{The histogram of the statistical information of textual descriptions. The word count of (a) denotes the length of textual descriptions. The word count of (b) denotes the number of words whose TF-IDF values are larger than 0.3.
Here we have removed stop-words.}
\label{fig:data_stat}
\end{center}
\end{figure}

\begin{table}[!htbp]
\centering
\begin{tabular}{cc|cc}
\toprule
\bf{Dataset} & \bf{Word Emb.} & \,\,\,\,\,\,\,\bf{MRR} & \bf{Hits@10} \\
\midrule
\textbf{NELL-ZS} & BERT Emb. &  \,\,\,\,\,\,\,0.237& \,\,\,\,35.5\,\,\,\, \\
 & \emph{Word2Vev} &  \,\,\,\,\,\,\,\textbf{0.253} & \textbf{37.1} \\
\textbf{Wiki-ZS} & BERT Emb. &  \,\,\,\,\,\,\,0.175 & 26.1 \\
 & \emph{Word2Vec} &  \,\,\,\,\,\,\,\textbf{0.208} & \textbf{29.4} \\
\bottomrule
\end{tabular}
\caption{\label{table:vsBERT} Link prediction comparison results between \emph{Word2Vec} and BERT embeddings as word representations.}
\end{table}

\subsection{Analysis of Textual Representations}
Figure \ref{fig:data_stat} illustrates the statistical information of text descriptions for two datasets. On the whole, the textual descriptions of NELL-ZS are longer than Wiki-ZS. However, after calculating their TF-IDF values, the number of highly-weighted words of both datasets are located in [2, 5]. 
For example, the highly-weighted words of relation \textsc{Worker} is \emph{livelihood}, \emph{employed} and \emph{earning}. It demonstrates the capacity of noise suppression. 
As for word representations\footnote{{ZSGAN}$_{KG}$ is not limited to a particular type of word embedding.}, besides \emph{Word2Vec}, we also attempt the contextualized word representations from BERT\footnote{We use the \emph{uncased-BERT-Base} model of hidden size 768.} \citep{devlin2019bert} as in Table \ref{table:vsBERT}. But their performance is less than satisfactory for two reasons: their high dimension and the sequence-level information involved in the representations. It is difficult for the generator to reduce dimension and extract discriminative features; So, GANs is hard to reach Nash equilibrium.

\subsection{Quality of Generated Data}
In Table \ref{table:per_relation}, we analyze the quality of the generated relation embeddings by our generator and present the comparable results of different relations against the ZS-DistMult, since ZS-DistMult is the best baseline model from Table \ref{table:results}. Unlike image, our generated data cannot be observed intuitively. Instead, we calculate the cosine similarity between the generated relation embeddings and the cluster center $x_c^r$ of their corresponding relations. It can be seen that our method indeed generates the plausible relation embeddings for many relations and the link prediction performance is positively correlated with the quality of the relation embeddings.

\subsection{Discussion}
In the respect of text information, we adopt the simple bag-of-words model rather than the neural-network-based text encoder, such CNN and LSTM. We indeed have tried these relatively complicated encoders, but their performance is barely satisfactory. We analyze that one of the main reasons is that the additional trainable parameter set involved in these encoders reduces the difficulty of adversarial training. In other words, the generator is more likely to over-fit the training set; Therefore, the generalization ability of generator is poor when dealing with unseen relations. Even though the bag-of-words model achieves better performance here, it still has the shortage of semantic diversity, especially when the understanding of a relation type needs consider the word sequence information in its textual description. In addition, as mentioned in the background, our zero-shot setting is based on an unified entity set $E$. It can be understood as expanding the current large-scale knowledge graph by adding the unseen relation edges between the existing entity nodes. It must be more beneficial to further consider the unseen entities. We leave these two points in future work.

\section{Conclusion}

In this paper, we propose a novel generative adversarial approach for zero-shot knowledge graph relational learning. We leverage GANs to generate plausible relation embeddings from raw textual descriptions. Under this condition, zero-shot learning is converted to the traditional supervised classification problem. An important aspect of our work is that our framework does not depend on the specific KG embedding methods, meaning that it is model-agnostic that could be potentially applied to any version of KG embeddings. Experimentally, our model achieves consistent improvements over various baselines on various datasets.

\section{Acknowledgements}
Pengda Qin is supported by China Scholarship Council and National Natural Science Foundation of China (61702047). Chunyun Zhang is supported by National Natural Science Foundation of China (61703234). Weiran Xu is supported by State Education Ministry -- China Mobile Research Fund Project (MCM20190701), DOCOMO Beijing Communications Laboratories Co., Ltd, National Key Research and Development Project No. 2019YFF0303302. The authors from UCSB are not supported by any of the projects above.

\bibliography{AAAI-QinP.7793}
\bibliographystyle{aaai}

\end{document}